\pdfoutput=1

\documentclass{article}

\usepackage{nips15submit_e}

\usepackage{url}            %
\usepackage{booktabs}       %

\usepackage{footnote}
\usepackage{graphicx}

\newcommand*\rot{\rotatebox{90}}

\title{High-performance Semantic Segmentation Using Very Deep Fully Convolutional Networks\thanks{This
    research was in part supported by the Data to Decisions
    Cooperative Research Centre.
    C. Shen's participation was in part
    supported by an ARC Future Fellowship (FT120100969).
    C. Shen is the corresponding author.
  }
}

\author{
  Zifeng Wu, Chunhua Shen, Anton van den Hengel\\
  The University of Adelaide,
  SA 5005, Australia\\
  \texttt{firstname.lastname@adelaide.edu.au}\\
}

\nipsfinalcopy %

\begin{document}

\maketitle

\begin{abstract}

We propose a method for high-performance semantic image segmentation (or semantic pixel
labelling) based on very deep residual networks,
which achieves the state-of-the-art performance.
A few design factors are carefully considered to this end.

We make the following contributions.
(i) First,  we evaluate  different variations of a fully convolutional residual network so as to find the best configuration,
including the number of layers, the resolution of feature maps, and the size of field-of-view.
Our experiments show that further enlarging the field-of-view and increasing the resolution of feature maps are typically
beneficial, which however inevitably leads to a higher demand for GPU memories.
To walk around the limitation, we propose a new method to simulate a high resolution network with a low resolution network,
which can be applied during training and/or testing.
(ii) Second, we propose an online bootstrapping method for training. We demonstrate that online bootstrapping is critically important
for achieving good accuracy.
(iii) Third we apply the traditional dropout to some of the residual blocks, which further improves the performance.
(iv) Finally, our method achieves the currently best mean intersection-over-union 78.3\% on the PASCAL VOC 2012 dataset,
as well as on the recent dataset Cityscapes.
\end{abstract}

\newpage

\section{Introduction}

Semantic image segmentation amounts to predicting the category of individual pixels in an image, which has been one of the
most active topics in the field of image understanding and computer vision for a long time.
Most of the recently proposed approaches to this task  are based on deep convolutional networks.
Particularly, the fully convolutional network (FCN)~\cite{FCN.CVPR.2015.Long} is
efficient and at the same time has achieved the state-of-the-art performance.
By reusing the computed feature maps for an image, FCN avoids redundant  re-computation for classifying
individual pixels in the image.
FCN  becomes the defacto  approach to dense prediction
and  methods  were proposed to further improve this framework, e.g., the DeepLab~\cite{DeepLab.ICLR.2015.Chen},
and the Adelaide-Context model \cite{AdelaideContext.2016.Lin}.
One key reason for the success of these methods is that they are based on
rich features learned from the very large ImageNet~\cite{ILSVRC2012} dataset, often in the form of a 16-layer VGGNet~\cite{VGGNet.2014.Simonyan}.
However, currently, there exist much improved models for image classification, e.g.,
the ResNet~\cite{ResNet.CVPR.2016.He,ResNet2.2016.He}.

To the best of our knowledge, building FCNs using ResNets is still an open topic to study on.
These networks are so deep that we would inevitably be faced with the limitation of GPU memories.
Besides, there are some aspects of the framework of FCN, which need to be explored carefully, such as
the size of field-of-view~\cite{DeepLab.ICLR.2015.Chen}, the resolution of feature maps, and the sampling strategies during training.
Based on the above consideration, here we attempt to fulfill the missing part of this topic.

In summary, we highlight the main contributions of this work
as follows:
\begin{itemize}
  \item
  We extensively evaluate  different variations of a fully convolutional residual network so as to find the best configuration,
including the number of layers, the resolution of feature maps, and the size of field-of-view.

We empirically demonstrate that  enlarging the field-of-view and increasing the resolution of feature maps are in general
beneficial. However, this  inevitably leads to a higher demand for GPU memories.
To solve this difficulty, we propose a new method to simulate a high resolution network with a low resolution network,
which can be applied during training and/or testing.

\item We propose an online bootstrapping method for training. We show that online bootstrapping is critically important
in achieving the best performance.

\item We apply dropout regularisation to some of the residual blocks, which further improves the performance.
Our method achieves the currently best results on VOC and Cityscapes datasets.
We achieve an mean intersection-over-union score on VOC of 78.3\% on the PASCAL VOC 2012,
which is a new record\footnote{\url{http://host.robots.ox.ac.uk:8080/anonymous/W382GA.html}
}.
\end{itemize}

\section{Related work}

In this section we briefly review the recent development of research on three topics, which are closely related to this paper.

\textbf{Very deep convolutional networks.}
The recent boom of deep convolution networks was originated when Krizhevsky~et~al.~\cite{CNN.NIPS.2012.Krizhevsky} won the first place in the ILSVRC 2012 competition~\cite{ILSVRC2012} with the 8-layer AlexNet.
The next year's  method `Clarifai'~\cite{ILSVRC2013} still had the same number of layers.
However, in 2014, the VGGNets~\cite{VGGNet.2014.Simonyan} were composed of up to nineteen layers,
while the even deeper 22-layer GoogLeNet~\cite{GoogLeNet.2014.Szegedy}  won the competition~\cite{ILSVRC2015}.
In 2015, the much deeper ResNets~\cite{ResNet.CVPR.2016.He} achieved the best performance~\cite{ILSVRC2015},
showing deeper networks indeed learn better features.
Nevertheless, the most impressive part was that He~et~al.~\cite{ResNet.CVPR.2016.He}
won in the object detection task with an overwhelming margin,
by replacing the VGGNets in Fast RCNN~\cite{FastRCNN.ICCV.2015.Girshick} with their ResNets,
which shows the importance of features in image understanding tasks.

The main  contribution that enables    them to
train so deep networks is that they connect some of the layers with shortcuts,
which directly pass through the signals and can thus avoid the vanishing gradient
effect which may be a problem for very deep plain networks.
In a more recent work, they redesigned their residual blocks to avoid over-fitting,
which enabled them to train an even deeper 200-layer residual network.
Deep ResNets can be seen as a simplified version of the highway network \cite{Highway}.

\textbf{Fully convolutional networks for semantic segmentation.}
Long~et~al.~\cite{FCN.CVPR.2015.Long} first proposed the framework of FCN for semantic segmentation, which is both effective and efficient.
They also enhanced the final feature maps with those from intermediate layers, which enables
their model to make finer predictions.
Chen~et~al.~\cite{DeepLab.ICLR.2015.Chen}  increased the resolution of feature maps by spontaneously
removing some of the down-sampling operations and accordingly introducing kernel dilation into their networks.
They also found that a classifier composed of small kernels with a large dilation performed as well as a classifier
with large kernels, and that reducing the size of field-of-view had an adverse impact on performance.
As post-processing, they applied dense CRFs to refine the predicted category score maps for further improvement.

Zheng~et~al.~\cite{CRFasRNN.ICCV.2015.Zheng} simulate   the dense CRFs with an
recurrent neural network
(RNN), which can be trained end-to-end together with the down-lying convolution layers.
Lin~et~al.~\cite{AdelaideContext.2016.Lin} jointly trained CRFs with down-lying convolution layers,
thus they are able to capture both `patch-patch' and `patch-background'
context with CRFs, rather than just pursue local smoothness as most of the previous methods do.

\textbf{Online bootstrapping for training deep convolutional networks.}
There are some recent works in the literature exploring sampling methods during training, which are concurrent with ours.
Loshchilov and Hutter~\cite{OHEM.ICLR.2016.Loshchilov} studied mini-batch selection in terms of image classification.
They picked hard training images from the whole training set according to their current losses, which were lazily updated once an image had been forwarded through the network being trained.
Shrivastava~et~al.~\cite{DetOHEM.CVPR.2016.Shrivastava} proposed to select hard region-of-interests (RoIs) for object detection.
They only computed the feature maps of an image once, and forwarded all RoIs of the image on top of these feature maps.
Thus they are able to find the hard RoIs  with a small extra computational cost.

The  method of  \cite{OHEM.ICLR.2016.Loshchilov}
is similar to ours in the sense that they all select hard training samples based on the current losses of individual data-points.
However, we only search hard pixels within the current mini-batch, rather than the whole training set.
In this sense, the  method of  \cite{DetOHEM.CVPR.2016.Shrivastava} is more similar to ours.
To our knowledge,  our method is  the first to propose online bootstrapping of hard
pixel samples for the problem of semantic image segmentation.

\section{Our method}

We first explain how to construct our baseline
fully convolutional residual network (FCRN) based on existing works in the literature,
mainly, the fully convolutional network (FCN)~\cite{FCN.CVPR.2015.Long} and the ResNet~\cite{ResNet.CVPR.2016.He}.
Then, we demonstrate how we can
walk around the limitation on GPU memories when training a very large network,
and finally introduce our method that applies online bootstrapping.

\subsection{Fully convolutional residual network}

We initialize a fully convolutional residual network from the original version of ResNet~\cite{ResNet.CVPR.2016.He} but not the newly proposed full pre-activation version~\cite{ResNet2.2016.He}.
From an original ResNet, we replace the linear classification layer with a convolution layer so as to make one prediction per spatial
location.
 Besides, we also remove the 7$\times$7 pooling layer.
This layer can enlarge the field-of-view (FoV)~\cite{DeepLab.ICLR.2015.Chen} of features, which is sometimes useful considering the fact that we human usually tell the category of a pixel by referring to its surrounding context region.
However, this pooling layer at the same time  smoothes the features.
In pixel labeling tasks, features of adjacent pixels should be distinct from each other when they respectively
belong to different categories, which may conflict with the pooling layer.
Therefore we  remove this layer and let the linear convolution layer on top deal with the FoV.

By now, the feature maps below the added linear convolution layer only has a resolution of 1/32,
which is apparently too low to precisely discriminate individual pixels.
Long~et~al.~\cite{FCN.CVPR.2015.Long} learned extra up-sampling layers to deal with this problem.
However, Chen~et~al.~\cite{DeepLab.ICLR.2015.Chen} reported that
the hole algorithm (or the \`{a}trous algorithm by Mallat~\cite{WaveletTour.2008.Mallat})  can be  more efficient.
Intuitively, the hole algorithm can be seen as dilating the convolution kernels before applying them to their input feature maps.
With this technique, we can build up a new network generating feature maps of any higher resolution, without changing the weights.
When there is a layer with down-sampling, we
skip the down-sampling part and increase the dilations of subsequent convolution kernels accordingly.
Refer to DeepLab~\cite{DeepLab.ICLR.2015.Chen} for a graphical explanation.

A sufficiently  large FoV was reported to be important by Chen~et~al.~\cite{DeepLab.ICLR.2015.Chen}.
Intuitively, we need to present  context information of a pixel to the top-most classification layer.
However, the features at different locations should be discriminative at the same time so that the classifier
can tell the differences between adjacent pixels which belong to different categories.
Therefore, a natural way is to let the classifier to handle the FoV, which can be achieved by enlarging its kernel size.
Unfortunately, the required size can be so large that it can  blow up the number of parameters in the classifier.
Nevertheless, we can resort to the hole algorithm again.
Thus we can use small kernels with large dilations in order to realize a large FoV.

In summary, following the above three steps, we design  the  baseline  FCRN.
Although the ResNet has shown its advantages in terms of many tasks due to much richer learned features,
we observe that  our baseline FCRN  is not  powerful enough to beat the best algorithm for
semantic segmentation~\cite{AdelaideContext.2016.Lin},
which is based on the VGGNet~\cite{VGGNet.2014.Simonyan}.

\subsection{Training of a large network with limited GPU memories}
The limitation of GPU memories is
one of the key problems during training of an FCN, and as well as an FCRN.
There are  at least two reasons to use more memories during training.

\textbf{To enlarge the FoV}.
It was reported by Chen~et~al.~\cite{DeepLab.ICLR.2015.Chen} that reducing the size of FoV from 224 down to 128 has an adverse impact on the performance of an FCN in terms of semantic segmentation.
What is more, we find that 224 is yet smaller than the optimal size.
To support an even larger FoV, we have to feed a network with larger input images, which may fire the limitation on GPU memories.

\textbf{To train with a high resolution}.
Many of the previous works~\cite{FCN.CVPR.2015.Long,DeepLab.ICLR.2015.Chen}
made predictions on top of feature maps with a resolution of either 1/16 or 1/8.
However, we find that a finer resolution of 1/4 can further improve the performance.
More importantly, although the models trained with different resolutions have the same number of parameters,
we can usually obtain a better model by training with a higher resolution.
Let $a$ be a network trained with a resolution of 1/16, while $b$ be trained with 1/8.
Intuitively, we anticipate that $b$ would outperform $a$, which is usually true since $b$ makes predictions at a higher resolution.
This comparison seems not that fair for $a$.
Therefore, we also test $a$ at a resolution of 1/8.
Nevertheless, the margin between $a$ and $b$ usually cannot be completely removed, according to our experiments.
In this sense, $b$ is still better than $a$ in the fairer comparison.
But unfortunately, increasing the resolution from 1/8 to 1/4 leads to four times larger feature maps, which may well exceed
current available GPU memories.

To this end, we modify the implementation of batch normalization~\cite{BatchNorm.2015.Ioffe}
in Caffe to apply a more conservative strategy in using GPU memories.
Then, we follow He~et~al.~\cite{ResNet.CVPR.2016.He} to fix the means and variances in all batch normalization layers,
which turns them into simple linear transformations.
Third, we reduce the number of images per mini-batch, which shows no adverse impact in our preliminarily experiments.
However, with these modifications, it is still not  feasible    to train a very deep FCRN with both large FoV and high resolutions.

One trivial approach is to feed a model with multiple small crops of the same image one by one,
and do not update the weights until gradients of all the crops have been aggregated.
However, there is still a compromise between large FoV and high resolution in this method.
With larger crops (to ensure large FoV), we will have to lower the resolution of feature maps.
On the other hand, with higher resolutions, we will have to reduce the size of each crop.

\begin{figure}[t!]
\centering
\includegraphics[width=0.95\linewidth,trim=0 340 0 0]{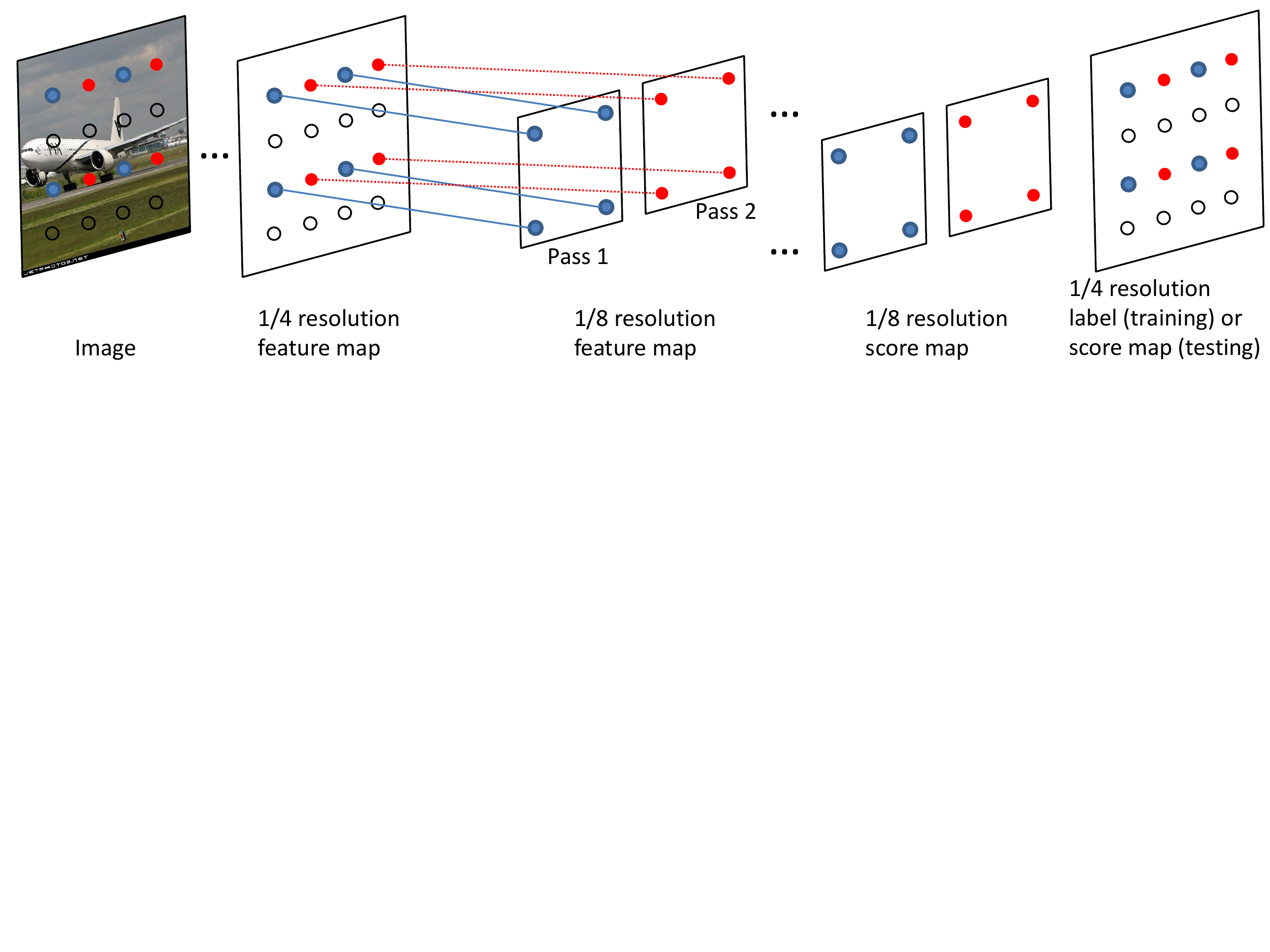}
\caption{Simulating  a high resolution model with a low resolution model.}
\label{fig:simulate_high_resolution}
\end{figure}

To break  this  dilemma,  we show how to simulate a high resolution model with a low resolution model.
We show an example in Fig.~\ref{fig:simulate_high_resolution}.
Suppose that we can indeed train a network whose score map resolution is 1/8 of the original input images,
while we are not able to train a 1/4 resolution one due to limited GPU memories.

So we resort to the 1/8 resolution model.
In the first pass, we feed the model with an image, which is large.
The feature maps will be down-sampled to 1/8 resolution at some intermediate layer, as depicted by the solid blue lines.
Naturally, the predicted score maps will also be at a resolution of 1/8.
During training, we  only  compute the loss and back-forward the gradients, but do not update the weights yet.
Here starts the second pass.
This time, before down-sampling, we first shift the 1/4 resolution feature maps horizontally
with a stride of one, so that the obtained 1/8 resolution feature maps are different from those in the first pass.
We do not update the weights until we finish the third and forth passes.
During testing, the idea is similar.
We only  put the obtained scores in four passes into their corresponding locations on 1/4 resolution scores maps.

\subsection{Online bootstrapping of hard training pixels}
When we train an FCN, depending on the size of image crops, there may be thousands of labeled pixels to predict per crop.
However, sometimes many of them can easily be discriminated from others, especially for those lying at the central part of a large semantic region.
Keeping on learning from these pixels can hardly improve the objective of semantic segmentation.
Based on the above consideration, we propose an online bootstrapping method, which forces networks to
focus on hard (and so more valuable) pixels during training.

Let there be $K$ different categories $c_j$ in a label space.
For simplicity, suppose that there is only one image crop per mini-batch, and let there be $N$ pixels $a_i$ to predict in this crop.
Let $y_i$ denote the ground truth label of pixel $a_i$, and $p_{ij}$ denote the predicted probability of pixel $a_i$ belonging to category $c_j$.
Then, the loss function can be defined as,
\begin{equation}
\ell = - \frac{1}{ \sum_{i}^{N} \sum_{j}^{K} 1\{y_i = j \textrm{ and } p_{ij} < t\} } ( \sum_{i}^{N} \sum_{j}^{K} 1\{y_i = j \textrm{ and } p_{ij} < t\} \log p_{ij})
\end{equation}
where $t \in (0, 1]$ is a threshold. Here
$1\{\cdot\}$ equals one when the condition inside holds, and otherwise equals zero.
In practice, we hope that there should be at least a reasonable number of pixels kept per mini-batch.
Hence, we will increase the threshold $t$ accordingly if the current model performs pretty well on a specific mini-batch.

\section{Experiments}

\subsection{Datasets}
We evaluate our method using two widely-used challenging datasets, i.e., the PASCAL VOC 2012~\cite{PascalVoc.IJCV.2014.Everingham} and the Cityscapes~\cite{Cityscapes.CVPR.2016.Cordts} datasets.

The PASCAL VOC 2012 dataset for semantic segmentation consists of photos taken in human daily life.
Besides the background category, there are twenty semantic categories to be predicted, including bus, car, cat, sofa, monitor, etc.
There are 1,464 fully labeled images for training (the train set) and another 1,449 for validating (the val set).
The ground-truth labels of the 1,456 images for testing (the test set) are not public, but there is an online evaluation server.
Following the conventional setting in the literature~\cite{FCN.CVPR.2015.Long,DeepLab.ICLR.2015.Chen}, we augment the train set with extra labeled PASCAL VOC images from the semantic boundaries dataset~\cite{SBD.ICCV.2011.Hariharan}.
So, in total there will be 10,582 for training.
The side lengths of images in this dataset are always no larger than 500 pixels.

The Cityscapes dataset consists of street scene images taken by car-carried cameras.
There are nineteen semantic categories to be predicted, including road, car, pedestrian, bicycle, etc.
There are 2975 fully labeled images for training (the train set) and another 500 for validating (the val set).
The ground-truth labels of images for testing (the test set) are not public, but there is an online evaluation server.
All of the images in this dataset are in the same size.
They are 1024 pixels high and 2048 pixels wide.

For evaluation, we  report:

(1) the pixel accuracy, which is the percentage of correctly labeled pixels on a whole test set;

(2) the mean pixel accuracy, which is the mean of class-wise pixel accuracies, and

(3) the mean class-wise intersection over union (IoU) scores.Note that we only show these three scores when it is possible for the individual datasets.
For example, only the mean IoU is available for the test set of PASCAL VOC 2012.

\subsection{Implementation details}
We implement our method based on Caffe~\cite{Caffe.2014.Jia},
and initialize fully convolutional residual networks~(FCRN) with the ResNet-50, ResNet-101 and ResNet-152 released by He~et~al.~\cite{ResNet.CVPR.2016.He}.
We evaluate the hyper-parameters of SGD using the validation sets of PASCAL VOC 2012 and Cityscapes.
We also apply random resizing and cropping to the original images to augment the training data.

\subsection{Results of the  vanilla FCRN}
In this subsection we investigate the impact of several configurations on the performance of a vanilla
FCRN, which include the network depth, the resolution of feature maps, the kernel size and dilation of the top-most classifier in the FCRN.

We show results on the val set of PASCAL VOC 2012 in Table~\ref{tbl:fcrn voc val}.
Firstly, we can achieve a significant improvement by increasing the depth from 50 to 101.
However, we observe clear over-fitting when increasing the depth to 152.

Secondly, generating feature maps with a higher resolution is also helpful.
Unfortunately, it is not easy to further increasing the resolution due to the limitation on GPU memories.
Thirdly, further increasing the size of FoV up to more than 224 is beneficial, which allows a classifier to learn from a larger context region surrounding a pixel.
However, note that all of the images in this dataset are no larger than 500$\times$500, and we feed a network with original images (without resizing) during testing.

Thus, we have to limit the size of FoV below 500 pixels on this dataset.
Otherwise, the dilated kernels of a classifier will be larger than the size of feature maps.
As a result, part of a kernel will be applied to padded zeros, which has no merit.
Similarly, if the size of FoV is larger than the size of image crops during training, part of a kernel cannot be properly learned.
In Table~\ref{tbl:fcrn voc val}, the largest FoV is 392.
No matter what is the depth, networks with this setting always achieve the best performance.
To realize such a large FoV, we can either enlarge the kernel size of the classifier or increase the dilation of these kernels.
However, this dilation should not be too large, since the feature vector per location can only cover a limited size of area.
For example, models with a dilation of eighteen show no obvious advantages over those with a dilation of twelve.
Especially, when the depth is 152, the model with a dilation of eighteen performs worse than the one with twelve.

\begin{table}[t]
\caption{Results of our vanilla FCRNs on the val set of PASCAL VOC 2012.}
\label{tbl:fcrn voc val}
\centering
\small
\begin{tabular}{ccccc|ccc}
\toprule
Depth & Resolution & Kernel & Dilation & FoV & Pixel acc.~\% & Mean acc.~\% & Mean IoU~\% \\
\hline\hline
50 & 1/16 & 3 & 6 & 208 & 92.74 & 78.68 & 69.09 \\
50 & 1/8 & 3 & 6 & 104 & 92.50 & 77.60 & 67.61 \\
50 & 1/8 & 3 & 12 & 200 & 93.03 & 79.51 & 69.94 \\
50 & 1/8 & 3 & 18 & 296 & 93.02 & 79.28 & 70.01 \\
50 & 1/8 & 5 & 6 & 200 & 92.98 & 79.34 & 69.81 \\
50 & \textbf{1/8} & \textbf{5} & \textbf{12} & \textbf{392} & \textbf{93.25} & \textbf{79.84} & \textbf{71.10} \\
50 & 1/8 & 7 & 6 & 296 & 93.14 & 79.54 & 70.67 \\
\hline\hline
101 & 1/16 & 3 & 6 & 208 & 93.22 & 80.16 & 70.93 \\
101 & 1/8 & 3 & 6 & 104 & 93.20 & 79.87 & 70.20 \\
101 & 1/8 & 3 & 12 & 200 & 93.68 & 81.29 & 72.34 \\
101 & 1/8 & 3 & 18 & 296 & 93.67 & 81.15 & 72.37 \\
101 & 1/8 & 5 & 6 & 200 & 93.52 & 81.00 & 71.97 \\
101 & \textbf{1/8} & \textbf{5} & \textbf{12} & \textbf{392} & \textbf{93.87} & \textbf{81.87} & \textbf{73.41} \\
101 & 1/8 & 7 & 6 & 296 & 93.61 & 81.34 & 72.56 \\
\hline\hline
152 & 1/16 & 3 & 6 & 208 & 93.32 & 80.21 & 71.27 \\
152 & 1/8 & 3 & 6 & 104 & 93.25 & 79.78 & 70.23 \\
152 & 1/8 & 3 & 12 & 200 & 93.65 & 80.97 & 72.29 \\
152 & 1/8 & 3 & 18 & 296 & 93.62 & 80.44 & 71.85 \\
152 & 1/8 & 5 & 6 & 200 & 93.59 & 80.92 & 71.95 \\
152 & \textbf{1/8} & \textbf{5} & \textbf{12} & \textbf{392} & \textbf{93.85} & \textbf{81.53} & \textbf{73.32} \\
152 & 1/8 & 7 & 6 & 296 & 93.81 & 81.38 & 72.94 \\
\bottomrule
\end{tabular}
\end{table}

We then show results on the val set of Cityscapes in Table~\ref{tbl:fcrn cityscapes val}.
Most of the observations on this dataset are consistent with those on PASCAL VOC 2012, as demonstrated above.
Two notable exceptions are as follows.
First, the problem of over-fitting seems lighter.
One possible reason is that the resolution of images in this dataset are higher than those in PASCAL VOC 2012, so the total number of pixels are actually larger.
On the other hand, the diversity of images in this dataset is smaller than those in PASCAL VOC 2012.
In this sense, even less training data can cover a larger proportion of possible situations, which can reduce over-fitting.
Second, 392 is still smaller than the optimal size of FoV.
Since the original images are in a size of 1024$\times$2048, we can feed a 50-layer network with larger image crops during both training and testing.
In this case, a network will prefer even larger FoV.
Therefore, to some extent, the ideal size of FoV depends on the size of image crops during training and testing.

\begin{table}[t]
\caption{Results of our vanilla FCRNs on the val set of Cityscapes.}
\label{tbl:fcrn cityscapes val}
\centering
\small
\begin{tabular}{ccccc|ccc}
\toprule
Depth & Resolution & Kernel & Dilation & FoV & Pixel acc.~\% & Mean acc.~\% & Mean IoU~\% \\
\hline\hline
50 & 1/16 & 3 & 6 & 208 & 93.83 & 74.67 & 66.41 \\
50 & 1/8 & 3 & 6 & 104 & 94.38 & 74.89 & 66.58 \\
50 & 1/8 & 3 & 12 & 200 & 94.47 & 75.91 & 67.68 \\
50 & 1/8 & 3 & 18 & 296 & 94.53 & 76.52 & 68.38 \\
50 & 1/8 & 5 & 6 & 200 & 94.48 & 76.17 & 68.04 \\
50 & 1/8 & 5 & 12 & 392 & 94.61 & 76.68 & 68.71 \\
50 & 1/8 & 5 & 18 & 584 & \textbf{94.64} & 76.34 & 68.53 \\
50 & 1/8 & 7 & 6 & 296 & 94.58 & \textbf{76.88} & \textbf{68.79} \\
50 & 1/8 & 7 & 12 & 584 & \textbf{94.64} & 76.57 & \textbf{68.79} \\
\hline\hline
101 & 1/16 & 3 & 6 & 208 & 94.11 & 76.26 & 67.62 \\
101 & 1/8 & 3 & 6 & 104 & 94.68 & 77.15 & 68.58 \\
101 & 1/8 & 3 & 12 & 200 & 94.78 & 78.30 & 69.99 \\
101 & 1/8 & 3 & 18 & 296 & 94.82 & 78.21 & 70.00 \\
101 & 1/8 & 5 & 6 & 200 & 94.75 & 78.11 & 69.89 \\
101 & \textbf{1/8} & \textbf{5} & \textbf{12} & \textbf{392} & \textbf{94.87} & \textbf{79.17} & \textbf{71.16} \\
101 & 1/8 & 7 & 6 & 296 & 94.75 & 78.43 & 70.40 \\
\hline\hline
152 & 1/16 & 3 & 6 & 208 & 94.26 & 76.89 & 68.30 \\
152 & 1/8 & 3 & 6 & 104 & 94.82 & 78.30 & 69.69 \\
152 & 1/8 & 3 & 12 & 200 & 94.94 & 78.79 & 70.66 \\
152 & 1/8 & 3 & 18 & 296 & 94.93 & 79.19 & 70.92 \\
152 & 1/8 & 5 & 6 & 200 & 94.88 & 78.77 & 70.61 \\
152 & \textbf{1/8} & \textbf{5} & \textbf{12} & \textbf{392} & \textbf{95.00} & \textbf{79.38} & \textbf{71.51} \\
152 & 1/8 & 7 & 6 & 296 & 94.91 & 79.08 & 70.87 \\
\bottomrule
\end{tabular}
\end{table}

\subsection{Impact of the feature map resolution}

In this subsection, we inspect the importance of training networks with a high resolution.
We only evaluate two 101-layer networks whose classifiers are composed of 5$\times$5 kernels, as shown in Table~\ref{tbl:fov and resolution}.
For each network, once we increase the resolution of predictions during testing, we consistently observe  a moderate improvement.

However, comparing the two networks at the same testing resolution, we find that the network trained with a resolution of 1/8 always performs better than the one trained with a resolution of 1/16.
As for the cause of this result, if we present finer labels to a network during training, we can force it to better discriminate the pixels located around semantic boundaries.
As the resolution increases, the labeled pixels for training become spatially closer, which makes them harder to discriminate.
However, a very deep network can learn from them anyway, and will probably perform better during testing.

\begin{table}[t]
\caption{Results showing the importance of training with a high resolution.}
\label{tbl:fov and resolution}
\centering
\small
\begin{tabular}{cc|ccc}
\toprule
Training resolution & Testing resolution & Pixel acc.~\% & Mean acc.~\% & Mean IoU~\% \\
\hline\hline
1/16 & 1/16 & 93.49 & 80.96 & 72.19 \\
1/16 & 1/8 & 93.75 & 81.71 & 72.96 \\
1/16 & 1/4 & 93.83 & 81.98 & 73.24 \\
\hline
1/8 & 1/8 & 93.87 & 81.87 & 73.41 \\
1/8 & 1/4 & 93.94 & 82.13 & 73.65 \\
\bottomrule
\end{tabular}
\end{table}

\subsection{Impact of online bootstrapping of hard training pixels}
In this subsection, we evaluate the impact of our proposed online bootstrapping.
We introduce this component into several representative FCRNs with  settings showing good performance
as evaluated previously, and test them on the PASCAL VOC 2012 and Cityscapes datasets.

The results are shown in Table~\ref{tbl:bootstrapping and dropout}.
In all cases, the best setting is to keep the 512 top hardest pixels.
The number of valid labels per image crop may be less than 512.
In this case, we keep all of them.
In spite of the consistence, we note that it actually depends on the size of image crops during training,
and these networks are trained with similar sizes of image crops.
When we increase the size of image crops, it will be better to keep more.
Otherwise we should keep less.

We also show the category-wise results in Tables~\ref{tbl:comparison voc}~and~\ref{tbl:comparison cityscapes}.
Generally speaking, the proposed bootstrapping can obviously improve the performance for those categories which are less frequent in training data, e.g., cow and horse on PASCAL VOC 2012, traffic light and train on Cityscapes.

Besides, to deal with the problem of over-fitting observed on the PASCAL VOC dataset, we introduce the traditional dropout~\cite{CNN.NIPS.2012.Krizhevsky} into some of the top-most blocks in FCRNs, which finally enables the 152-layer network to outperform the 101-layer network.

\begin{table}[t]
\caption{Results with online bootstrapping and/or traditional dropout.}
\label{tbl:bootstrapping and dropout}
\centering
\small
\begin{tabular}{ccccc|ccc}
\toprule
Depth & Resolution & Kernel & Dilation & Bs./Do. & Pixel acc.~\% & Mean acc.~\% & Mean IoU~\% \\
\hline\hline
\multicolumn{8}{c}{PASCAL VOC 2012} \\
\hline\hline
101 & 1/8 & 5 & 12 & F/F & 93.87 & 81.87 & 73.41 \\
\hline
101 & 1/8 & 5 & 12 & 256/F & 94.11 & 81.44 & 74.41 \\
101 & 1/8 & 5 & 12 & \textbf{512}/F & \textbf{94.23} & \textbf{82.09} & \textbf{74.80} \\
101 & 1/8 & 5 & 12 & 1024/F & 94.08 & 81.84 & 74.17 \\
\hline
101 & 1/8 & 5 & 12 & 512/T & 94.27 & 84.82 & 75.45 \\
\hline\hline
152 & 1/8 & 5 & 12 & F/F & 93.85 & 81.53 & 73.32 \\
\hline
152 & 1/8 & 5 & 12 & 256/F & 94.19 & 81.64 & 74.53 \\
152 & 1/8 & 5 & 12 & \textbf{512}/F & \textbf{94.23} & \textbf{82.06} & \textbf{74.72} \\
152 & 1/8 & 5 & 12 & 1024/F & 94.15 & 81.94 & 74.40 \\
\hline
152 & 1/8 & 5 & 12 & 512/T & 94.46 & 85.13 & 75.90 \\
\hline\hline
\multicolumn{8}{c}{Cityscapes} \\
\hline\hline
152 & 1/8 & 5 & 12 & F/F & 95.00 & 79.38 & 71.51 \\
\hline
152 & 1/8 & 5 & 12 & 256/F & 95.41 & 81.37 & 73.97 \\
152 & 1/8 & 5 & 12 & \textbf{512}/F & \textbf{95.46} & \textbf{82.04} & \textbf{74.64} \\
152 & 1/8 & 5 & 12 & 1024/F & 95.38 & 81.00 & 73.45 \\
\bottomrule
\end{tabular}
\end{table}

\subsection{Comparison with previous state-of-the-art}

We compare our method with the previous best performers on the PASCAL VOC 2012 datasets in Table~\ref{tbl:comparison voc}.
When training our model only with the PASCAL VOC data, we achieve a remarkable improvement in terms of mean IoU.
Our method outperforms the previous best performer by 2.0\% and wins the first place for twelve out of the twenty categories.

When pre-training our model with the Microsoft COCO~\cite{COCO.ECCV.2014.Lin} data,
we achieve a moderate improvement of 1.0\% and win the first place for thirteen out of the twenty categories.
Generally speaking, our method usually loses for those very hard categories, e.g., `bicycle', `chair', `diningtable', `pottedplant'
and `sofa', for which most of the methods can only achieve scores below 70.0\%.

The instances of these categories are usually of great diversity and in occluded situation, suggesting that more training data would be needed.
But unfortunately, they are generally the less frequent categories in the training data of PASCAL VOC 2012.

\begin{savenotes}
\begin{table}[t]
\caption{Category-wise and mean IoU scores on the PASCAL VOC 2012 dataset.}
\label{tbl:comparison voc}
\setlength{\tabcolsep}{1pt}
\centering
\footnotesize
\resizebox{\textwidth}{!}
{
\begin{tabular}{r|cccccccccccccccccccc|c}
  \toprule
\rot{Method} & \rot{aeroplane} & \rot{bicycle} & \rot{bird} & \rot{boat} & \rot{bottle} & \rot{bus} & \rot{car} & \rot{cat} & \rot{chair} & \rot{cow} & \rot{diningtable} & \rot{dog} & \rot{horse} & \rot{motorbike} & \rot{person} & \rot{pottedplant} & \rot{sheep} & \rot{sofa} & \rot{train} & \rot{tvmonitor} & \rot{Mean}   \\
\hline\hline
\multicolumn{22}{c}{Results on val set} \\
\hline
FCRN & 86.7 & 39.5 & 85.5 & 66.9 & 79.3 & 90.7 & 84.7 & 90.6 & 34.0 & 79.1 & 51.6 & 83.9 & 80.6 & 80.0 & 83.0 & 55.7 & 80.6 & 40.3 & 82.7 & 72.9 & 73.4 \\
\hline
FCRN + Bs. & 88.3 & 40.4 & 86.5 & 66.6 & 80.1 & 91.6 & 84.3 & 90.1 & 36.6 & 83.7 & 53.6 & 84.5 & 85.1 & 79.9 & 83.9 & 59.0 & 83.3 & 44.6 & 81.1 & 74.5 & 74.8 \\
\hline\hline
\multicolumn{22}{c}{Results on test set obtained with models trained only using PASCAL VOC data} \\
\hline
FCN-8s~\cite{FCN.CVPR.2015.Long} & 76.8 & 34.2 & 68.9 & 49.4 & 60.3 & 75.3 & 74.7 & 77.6 & 21.4 & 62.5 & 46.8 & 71.8 & 63.9 & 76.5 & 73.9 & 45.2 & 72.4 & 37.4 & 70.9 & 55.1 & 62.2 \\
DeepLab~\cite{DeepLab.ICLR.2015.Chen} & 84.4 & 54.5 & 81.5 & 63.6 & 65.9 & 85.1 & 79.1 & 83.4 & 30.7 & 74.1 & 59.8 & 79.0 & 76.1 & 83.2 & 80.8 & 59.7 & 82.2 & 50.4 & 73.1 & 63.7 & 71.6 \\
CRFasRNN~\cite{CRFasRNN.ICCV.2015.Zheng} & 87.5 & 39.0 & 79.7 & 64.2 & 68.3 & 87.6 & 80.8 & 84.4 & 30.4 & 78.2 & 60.4 & 80.5 & 77.8 & 83.1 & 80.6 & 59.5 & 82.8 & 47.8 & 78.3 & 67.1 & 72.0 \\
DeconvNet~\cite{DeconvNet.ICCV.2015.Noh} & 89.9 & 39.3 & 79.7 & 63.9 & 68.2 & 87.4 & 81.2 & 86.1 & 28.5 & 77.0 & 62.0 & 79.0 & 80.3 & 83.6 & 80.2 & 58.8 & 83.4 & 54.3 & 80.7 & 65.0 & 72.5 \\
DPN~\cite{DPN.ICCV.2015.Liu} & 87.7 & \textbf{59.4} & 78.4 & 64.9 & 70.3 & 89.3 & 83.5 & 86.1 & 31.7 & 79.9 & \textbf{62.6} & 81.9 & 80.0 & 83.5 & 82.3 & 60.5 & 83.2 & 53.4 & 77.9 & 65.0 & 74.1 \\
UoA-Context~\cite{AdelaideContext.2016.Lin} & 90.6 & 37.6 & 80.0 & \textbf{67.8} & 74.4 & \textbf{92.0} & 85.2 & 86.2 & \textbf{39.1} & 81.2 & 58.9 & 83.8 & 83.9 & 84.3 & 84.8 & \textbf{62.1} & 83.2 & \textbf{58.2} & 80.8 & \textbf{72.3} & 75.3 \\
\hline
ours & \textbf{91.9} & 46.5 & \textbf{91.2} & 65.1 & \textbf{74.8} & 90.9 & \textbf{87.2} & \textbf{94.9} & 32.2 & \textbf{88.7} & 57.5 & \textbf{90.5} & \textbf{90.9} & \textbf{87.8} & \textbf{86.0} & 60.8 & \textbf{88.5} & 52.3 & \textbf{84.5} & 67.9 & \textbf{77.3}\footnote{\url{http://host.robots.ox.ac.uk:8080/anonymous/4TBLCE.html}} \\
\hline\hline
\multicolumn{22}{c}{Results on test set obtained with models trained using PASCAL VOC + COCO data} \\
\hline
DeepLab~\cite{DeepLab.ICLR.2015.Chen} & 89.1 & 38.3 & 88.1 & 63.3 & 69.7 & 87.1 & 83.1 & 85.0 & 29.3 & 76.5 & 56.5 & 79.8 & 77.9 & 85.8 & 82.4 & 57.4 & 84.3 & 54.9 & 80.5 & 64.1 & 72.7 \\
CRFasRNN~\cite{CRFasRNN.ICCV.2015.Zheng} & 90.4 & 55.3 & \textbf{88.7} & 68.4 & 69.8 & 88.3 & 82.4 & 85.1 & 32.6 & 78.5 & 64.4 & 79.6 & 81.9 & 86.4 & 81.8 & 58.6 & 82.4 & 53.5 & 77.4 & 70.1 & 74.7 \\
DPN~\cite{DPN.ICCV.2015.Liu} & 89.0 & \textbf{61.6} & 87.7 & 66.8 & 74.7 & 91.2 & 84.3 & 87.6 & 36.5 & 86.3 & \textbf{66.1} & 84.4 & 87.8 & 85.6 & 85.4 & 63.6 & 87.3 & 61.3 & 79.4 & 66.4 & 77.5 \\
UoA-Context~\cite{AdelaideContext.2016.Lin} & \textbf{92.9} & 39.6 & 84.0 & 67.9 & 75.3 & \textbf{92.7} & 83.8 & 90.1 & \textbf{44.3} & 85.5 & 64.9 & 87.3 & 88.8 & 84.5 & 85.5 & \textbf{68.1} & 89.0 & \textbf{62.8} & 81.2 & 71.4 & 77.8 \\
\hline
ours & \textbf{92.9} & 45.6 & 88.1 & \textbf{70.1} & \textbf{75.4} & 90.9 & \textbf{88.7} & \textbf{94.3} & 35.2 & \textbf{86.8} & 60.6 & \textbf{89.7} & \textbf{91.6} & \textbf{88.5} & \textbf{87.0} & 62.6 & \textbf{89.4} & 55.1 & \textbf{85.4} & \textbf{72.4} & \textbf{78.3}\footnote{\url{http://host.robots.ox.ac.uk:8080/anonymous/W382GA.html}} \\
\bottomrule
\end{tabular}
}
\end{table}
\end{savenotes}

\begin{savenotes}
\begin{table}[t]
\caption{Category-wise and mean IoU scores on the Cityscapes dataset.}
\label{tbl:comparison cityscapes}
\setlength{\tabcolsep}{1pt}
\centering
\resizebox{\textwidth}{!}
{
\begin{tabular}{r|ccccccccccccccccccc|c}
\toprule
\rot{Method} & \rot{road} & \rot{sidewalk} & \rot{building} & \rot{wall} & \rot{fence} & \rot{pole} & \rot{traffic light} & \rot{traffic sign} & \rot{vegetation} & \rot{terrain} & \rot{sky} & \rot{person} & \rot{rider} & \rot{car} & \rot{truck} & \rot{bus} & \rot{train} & \rot{motorcycle} & \rot{bicycle} & \rot{Mean}
\\
\hline\hline
\multicolumn{21}{c}{Results on val set} \\
\hline
FCRN & 97.4 & 80.3 & 90.8 & 47.6 & 53.8 & 53.1 & 58.1 & 70.2 & 91.2 & 59.6 & 93.2 & 77.1 & 54.4 & 93.0 & 67.1 & 79.4 & 62.2 & 57.3 & 72.7 & 71.5 \\
\hline
FCRN + Bs. & 97.6 & 82.0 & 91.7 & 52.3 & 56.2 & 57.0 & 65.7 & 74.4 & 91.7 & 62.5 & 93.8 & 79.8 & 59.6 & 94.0 & 66.2 & 83.7 & 70.3 & 64.2 & 75.5 & 74.6 \\
\bottomrule
\end{tabular}
}
\end{table}
\end{savenotes}

\subsection{Discussions}
Several other variations we have evaluated are as follows.
The first is to set a larger learning rate for the newly added convolution layer, which shows no obvious advantage in most of our experiments.
This is not consistent with how we usually fine-tune a VGGNet, e.g., in DeepLab~\cite{DeepLab.ICLR.2015.Chen}.
There seems be some differences to be explored between tuning a residual network and a traditional network.
The second is to add random color noise to the images, just as Krizhevsky~et~al.~\cite{CNN.NIPS.2012.Krizhevsky} did, which shows no improvement either.
We have to add the same noise to a whole image crop, compared with adding 128 different noises per mini-batch~\cite{CNN.NIPS.2012.Krizhevsky}, which might be the reason why this data augmentation approach does not work in our experiments.
Besides, as mentioned before, we observe no obvious adverse impact when decreasing the number of images involved in one mini-batch.
An intuition is to use an enough large group of images per mini-batch, e.g., FCN~\cite{FCN.CVPR.2015.Long} and DeepLab~\cite{DeepLab.ICLR.2015.Chen} both used 20 per mini-batch.
However, according to our experiments, it is not that necessary for semantic segmentation as it does for image classification.

\section{Conclusions}

In this work, we have built  a  few  fully convolutional residual networks and explored their performances for the task of
semantic image segmentation.
We have shown the importance of large field-of-view and high resolution features maps.
To break the limitation of GPU memories, we have proposed to simulate a high resolution network with a low resolution network.
More importantly, we have proposed an online bootstrapping method  to mine hard training pixels, which
significantly  improve the accuracy.
Finally, we have achieved the state-of-the-art mean IoU score on the PASCAL VOC 2012 dataset.

\small
\newpage
\bibliographystyle{IEEEtran}
\bibliography{CSRef}

\begin{thebibliography}{10}
\providecommand{\url}[1]{#1}
\csname url@samestyle\endcsname
\providecommand{\newblock}{\relax}
\providecommand{\bibinfo}[2]{#2}
\providecommand{\BIBentrySTDinterwordspacing}{\spaceskip=0pt\relax}
\providecommand{\BIBentryALTinterwordstretchfactor}{4}
\providecommand{\BIBentryALTinterwordspacing}{\spaceskip=\fontdimen2\font plus
\BIBentryALTinterwordstretchfactor\fontdimen3\font minus
  \fontdimen4\font\relax}
\providecommand{\BIBforeignlanguage}[2]{{%
\expandafter\ifx\csname l@#1\endcsname\relax
\typeout{** WARNING: IEEEtran.bst: No hyphenation pattern has been}%
\typeout{** loaded for the language `#1'. Using the pattern for}%
\typeout{** the default language instead.}%
\else
\language=\csname l@#1\endcsname
\fi
#2}}
\providecommand{\BIBdecl}{\relax}
\BIBdecl

\bibitem{FCN.CVPR.2015.Long}
J.~Long, E.~Shelhamer, and T.~Darrell, ``Fully convolutional networks for
  semantic segmentation,'' in \emph{Proc. IEEE Conf. Comp. Vis. Patt. Recogn.},
  2015.

\bibitem{DeepLab.ICLR.2015.Chen}
L.~Chen, G.~Papandreou, I.~Kokkinos, K.~Murphy, and A.~Yuille, ``Semantic image
  segmentation with deep convolutional nets and fully connected {CRF}s,'' in
  \emph{Proc. Int. Conf. Learn. Representations}, 2015.

\bibitem{AdelaideContext.2016.Lin}
G.~Lin, C.~Shen, A.~van~den Hengel, and I.~Reid, ``Exploring context with deep
  structured models for semantic segmentation,'' arXiv:1603.03183, 2016.

\bibitem{ILSVRC2012}
J.~Deng, A.~Berg, S.~Satheesh, H.~Su, A.~Khosla, and L.~Fei-Fei, ``Image{N}et
  {L}arge {S}cale {V}isual {R}cognition {C}hallenge 2012 ({ILSVRC 2012}),''
  2012.

\bibitem{VGGNet.2014.Simonyan}
K.~Simonyan and A.~Zisserman, ``Very deep convolutional networks for
  large-scale image recognition,'' arXiv:1409.1556, 2014.

\bibitem{ResNet.CVPR.2016.He}
K.~He, X.~Zhang, S.~Ren, and J.~Sun, ``Deep residual learning for image
  recognition,'' in \emph{Proc. IEEE Conf. Comp. Vis. Patt. Recogn.}, 2016.

\bibitem{ResNet2.2016.He}
------, ``Identity mappings in deep residual networks,'' arXiv:1603.05027,
  2016.

\bibitem{CNN.NIPS.2012.Krizhevsky}
A.~Krizhevsky, I.~Sutskever, and G.~Hinton, ``Image{N}et classification with
  deep convolutional neural networks,'' in \emph{Proc. Advances in Neural Inf.
  Process. Syst.}, 2012.

\bibitem{ILSVRC2013}
O.~Russakovsky, J.~Deng, J.~Krause, A.~Berg, and L.~Fei-Fei, ``Image{N}et
  {L}arge {S}cale {V}isual {R}cognition {C}hallenge 2013 ({ILSVRC 2013}),''
  2013.

\bibitem{GoogLeNet.2014.Szegedy}
C.~Szegedy, W.~Liu, Y.~Jia, P.~Sermanet, S.~Reed, D.~Anguelov, D.~Erhan,
  V.~Vanhoucke, and A.~Robinovich, ``Going deeper with convolutions,''
  arXiv:1409.4842, 2014.

\bibitem{ILSVRC2015}
O.~Russakovsky, J.~Deng, H.~Su, J.~Krause, S.~Satheesh, S.~Ma, Z.~Huang,
  A.~Karpathy, A.~Khosla, M.~Bernstein, A.~Berg, and L.~Fei-Fei, ``Image{N}et
  {L}arge {S}cale {V}isual {R}cognition {C}hallenge,'' 2015.

\bibitem{FastRCNN.ICCV.2015.Girshick}
R.~Girshick, ``Fast {R}-{CNN},'' in \emph{Proc. IEEE Int. Conf. Comp. Vis.},
  2015.

\bibitem{Highway}
R.~K. Srivastava, K.~Greff, and J.~Schmidhuber, ``Training very deep
  networks,'' in \emph{Proc. Advances in Neural Inf. Process. Syst.}, 2015.

\bibitem{CRFasRNN.ICCV.2015.Zheng}
S.~Zheng, S.~Jayasumana, B.~Romera-Paredes, V.~Vineet, Z.~Su, D.~Du, C.~Huang,
  and P.~Torr, ``Conditional random fields as recurrent neural networks,'' in
  \emph{Proc. IEEE Int. Conf. Comp. Vis.}, 2015.

\bibitem{OHEM.ICLR.2016.Loshchilov}
I.~Loshchilov and F.~Hutter, ``Online batch selection for faster training of
  neural networks,'' in \emph{Proc. Int. Conf. Learn. Representations}, 2016.

\bibitem{DetOHEM.CVPR.2016.Shrivastava}
A.~Shrivastava, A.~Gupta, and R.~Girshick, ``Training region-based object
  detectors with online hard example mining,'' in \emph{Proc. IEEE Conf. Comp.
  Vis. Patt. Recogn.}, 2016.

\bibitem{WaveletTour.2008.Mallat}
S.~Mallat, \emph{A wavelet tour of signal processing}, 3rd~ed.\hskip 1em plus
  0.5em minus 0.4em\relax Academic Press, December 2008.

\bibitem{BatchNorm.2015.Ioffe}
S.~Ioffe and C.~Szegedy, ``Batch normalization: Accelerating deep network
  training by reducing internal covariate shift,'' arXiv:1502.03167, 2015.

\bibitem{PascalVoc.IJCV.2014.Everingham}
M.~Everingham, S.~Eslami, L.~van Gool, C.~Williams, J.~Winn, and A.~Zisserman,
  ``The {PASCAL} visual object classes challenge: {A} retrospective,''
  \emph{Int. J. Computer Vision}, 2014.

\bibitem{Cityscapes.CVPR.2016.Cordts}
M.~Cordts, M.~Omran, S.~Ramos, T.~Rehfeld, M.~Enzweiler, R.~Benenson,
  U.~Franke, S.~Roth, and B.~Schiele, ``The {C}ityscapes dataset for semantic
  urban scene understanding,'' in \emph{Proc. IEEE Conf. Comp. Vis. Patt.
  Recogn.}, 2016.

\bibitem{SBD.ICCV.2011.Hariharan}
B.~Hariharan, P.~Arbelaez, L.~Bourdev, S.~Maji, and J.~Malik, ``Semantic
  contours from inverse detectors,'' in \emph{Proc. IEEE Int. Conf. Comp.
  Vis.}, 2011.

\bibitem{Caffe.2014.Jia}
Y.~Jia, E.~Shelhamer, J.~Donahue, S.~Karayev, J.~Long, R.~Girshick,
  S.~Guadarrama, and T.~Darrell, ``Caffe: {C}onvolutional architecture for fast
  feature embedding,'' arXiv:1408.5093, 2014.

\bibitem{COCO.ECCV.2014.Lin}
T.~Lin, M.~Maire, S.~Belongie, J.~Hays, P.~Perona, D.~R.~P. Doll\'{a}r, and
  C.~Zitnick, ``Microsoft {COCO}: {C}ommon objects in context,'' in \emph{Proc.
  Eur. Conf. Comp. Vis.}, 2014.

\bibitem{DeconvNet.ICCV.2015.Noh}
H.~Noh, S.~Hong, and B.~Han, ``Learning deconvolution network for semantic
  segmentation,'' in \emph{Proc. IEEE Int. Conf. Comp. Vis.}, 2015.

\bibitem{DPN.ICCV.2015.Liu}
Z.~Liu, X.~Li, P.~Luo, C.~Loy, and X.~Tang, ``Semantic image segmentation via
  deep parsing network,'' in \emph{Proc. IEEE Int. Conf. Comp. Vis.}, 2015.

\end{thebibliography}

\end{document}